\pdfoutput=1

\documentclass[11pt]{article}

\usepackage{acl}

\usepackage{subcaption}
\usepackage{times}
\usepackage{tipa}
\usepackage{gb4e}
\noautomath
\usepackage{latexsym}
\usepackage{tikz}
\usepackage[utf8]{inputenc}

\usepackage{booktabs}

\usepackage[T1]{fontenc}

\usepackage[utf8]{inputenc}

\usepackage{microtype}

\usepackage{inconsolata}

\usepackage{graphicx}

\usepackage{colortbl}

\usepackage{amsmath}
\usepackage{pgfmath}

\newcommand{\maxdiff}{0.65}

\newcommand{\heatcell}[2]{%
  \pgfmathparse{abs(#1)}%
  \let\absval\pgfmathresult
  \pgfmathparse{int(min(max(\absval/\maxdiff*100,0),100))}%
  \let\perct\pgfmathresult
  \ifdim #2 pt<0.05pt
    \begingroup
      \edef\x{%
        \endgroup
        \noexpand\cellcolor{red!\perct!yellow}%
        \noexpand\textbf{#1}%
      }%
    \x
  \else
    #1%
  \fi
}

\title{Clarifying orthography: Orthographic transparency as compressibility}

\author{Charles J. Torres \\
  University of California, Irvine \\
  \texttt{\href{mailto:charlt4@uci.edu}{charlt4@uci.edu}} \\\And
  Richard Futrell \\
  University of California, Irvine \\
  \texttt{\href{mailto:rfutrell@uci.edu}{rfutrell@uci.edu}}}

\begin{document}
\maketitle
\begin{abstract}

Orthographic transparency---how directly spelling is related to sound---lacks a unified, script-agnostic metric. Using ideas from algorithmic information theory, we quantify orthographic transparency in terms of the mutual compressibility between orthographic and phonological strings. Our measure provides a principled way to combine two factors that decrease orthographic transparency, capturing both irregular spellings and rule complexity in one quantity. We estimate our transparency measure using prequential code-lengths derived from neural sequence models.  Evaluating 22 languages across a broad range of script types (alphabetic, abjad, abugida, syllabic, logographic) confirms common intuitions about relative transparency of scripts. Mutual compressibility offers a simple, principled, and general yardstick for orthographic transparency.

\vspace{.3em}
\hspace{1.25em}\includegraphics[width=1.25em,height=1.25em]{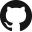}{\hspace{.75em}\parbox{\dimexpr\linewidth-2\fboxsep-2\fboxrule}{\url{https://github.com/cj-torres/opacity-measure}}}
\vspace{.1em}
\end{abstract}

\section{Introduction}

Orthographic transparency, conversely known as orthographic opacity or orthographic depth, refers to the extent that orthography encodes phonology in a way that is transparent or simple. For example, English readers will already be familiar with the sometimes poor correspondence English spelling has with its pronunciation, meaning English orthography is relatively opaque. Likewise, Chinese characters give very little---sometimes no---information about the pronunciation of their corresponding morphemes, being even more opaque than English. Spanish, on the other hand, is known for the extreme ease with which pronunciations can be extracted from spelling alone, making its orthography relatively transparent.

Despite the intuitive nature of the concept or orthographic transparency, it has no precise agreed-upon operational definition. Various measurements have attempted to associate orthographic transparency with numbers of grapheme-to-phoneme correspondence (GPC) rules, irregularity of the coverage of these rules \citep{ziegler2000drc}, or a combination of the two \citep{van1994measuring, protopapas2009greek}, or proxies for these such as onset entropy \citep{borgwaldt2005onset} which measures an unnormalized conditional Shannon entropy of the first phoneme given the first grapheme. 
Furthermore, existing measures rarely accommodate non-alphabetic systems \citep{borleffs2017measuring}, a striking limitation given the breadth of existing writing systems. In addition to its scientific interest, finding a transparency measurement has promise for real pedagogical applications, as it is commonly understood to impact the speed of reading acquisition \citep{aro2003learning, seymour2003foundation}. 

We offer a new method to quantify orthographic transparency using a measure which combines different sources of opacity in a principled manner. Using ideas from algorithmic information theory \citep{li2008introduction,shen2017kolmogorov} we posit that orthographic transparency can be measured as the shared structure between the orthography and phonology of a language. We show that this measure not only succeeds in unifying the aforementioned sources of opacity, but that it can generalize to any writing system (and, in fact, any two datatypes), and that it conforms to existing results and intuitions. In the remainder of the paper we will first provide background information on orthographic transparency and algorithmic information theory. After this background we will explain the methods we used to extract our measurements and present results a multilingual survey, and Japanese scripts in particular. then compare these to some existing metrics.



\begin{figure*}[h!]
    \centering
    \includegraphics[width=\textwidth]{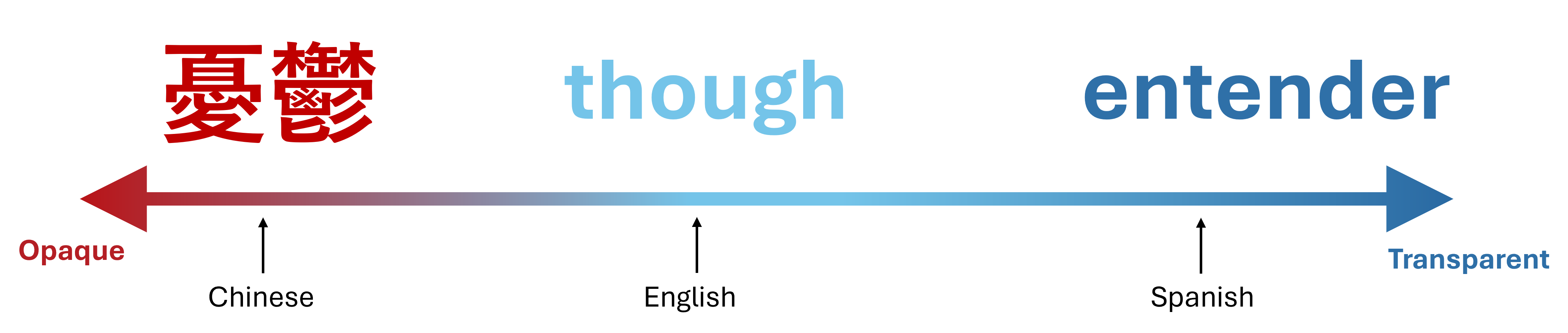}
    \caption{The hypothetical orthographic transparency continuum imagines languages and their writing systems organized on a one-dimensional line.}
    \label{fig:continuum}
\end{figure*}

\section{Background}
\label{sec:background}


The concept of orthographic transparency imagines that languages' writing systems can be arranged along a continuum (Figure~\ref{fig:continuum}). Languages with writing systems that are highly informative are imagined at one end of the continuum, and language's with writing systems that are highly uninformative about their pronunciation are at the other end. This simple classification belies different sources that contribute to informativity, however.

Consider the famous example of the multiple pronunciations of the suffix \texttt{-ough} in English orthography. There is no consistent rule that can govern all the transformations in \texttt{tough}, \texttt{through}, \texttt{cough}, and \texttt{dough} beyond memorizing the entire sequence of graphemes and how its corresponding word is pronounced. Even worse are homographs like \texttt{lead} for which contextual information must assist in determining the correct pronunciation. Such mappings are non-deterministic, and are characterized in the literature via irregularity measurements, by crafting rules---usually simple GPC rules---and measuring the coverage of those rules \citep{van1994measuring, protopapas2009greek} or more recently via training sequence-to-sequence networks and measuring performance on holdout sets \citep{marjou2019oteann}. This measurement can operate in both the orthography-to-phonology and phonology-to-orthography direction.

Contrast this source of opacity with another. French is often considered severely to moderately opaque in the literature \citep{seymour2003foundation, goswami1998children}. This is due to the complexity of the mappings and unpredictability of the spellings. Consider the French word \texttt{oiseau} pronounced /wazo/. This pronunciation is in fact derivable from the spelling via regular grapheme-to-phoneme correspondences. However, the correspondences are complex, mapping multiple graphemes to phonemes. This kind of opacity can be captured by counting the number of rules which can cover a language, procedures which have also been done \citep{van1994measuring}.

The existence of these two kinds of opacity (irregularity and complexity of mappings) have led some to outright reject the notion of orthographic opacity as a single concept at all \citep{schmalz2015getting}. If the concept of opacity is accepted as meaningful, then any measurement of it must employ some estimate of the relative importance of both sources to the total measurement, or find some means of unifying these two sources of opacity. One measurement, the \textit{onset entropy} \citep{borgwaldt2005onset}, does offer such a unification, but here it encounters a different weakness, and is only able to be used at the beginning of a word. 

Yet it seems desirable to find such a more general unification, as opacity is linked with reading proficiency during reading acquisition \citep{aro2003learning, seymour2003foundation}. We show tha such a unification is possible in a principled and practical way. We propose that a unification of the sources is natural: both of these sources of opacity contribute to making the orthography \emph{mutually incompressible} with the phonology of the language (and vice versa). Below we review what we mean by mutual compressibility, and how this can be measured for orthographic systems. Then we will demonstrate results showing that this measure is a natural fit for existing data.


\section{Algorithmic Information and Mutual Compressibility}

We propose to measure orthographic transparency notions of complexity and compressibility from the algorithmic information theory literature \citep{li2008introduction, shen2017kolmogorov}. Here we introduce and motivate our measure, and describe how it can be computed using neural sequence models.

Some definitions are in order before we introduce our measure. The first is the function $K(x)$. This measure, known as \textbf{algorithmic information} or Kolmogorov complexity, is the length of the shortest computer program which outputs the string $x$. Intuitively, this reflects the compressibility of the string $x$: a low value indicates that $x$ is highly regular---containing some sort of pattern, which can be represented with a short program. For example, consider the strings below:
\begin{exe}
    \ex\label{ex:repeating} $a=1010101010101010101010101...$
    \ex\label{ex:random} $b=1100001110101001100100101...$
\end{exe}
Example~(\ref{ex:repeating}) contains a regular pattern. The longer it becomes, the more it will benefit from this regularity under compression. In other words, the algorithmic information $K(a)$ is small. Example~(\ref{ex:random}) is randomly sampled from a Bernoulli distribution, so the algorithmic information $K(b)$ should be large. There is no pattern, and a longer random sequence won't be any more compressible.

With the introduction of algorithmic information comes a second measure. \textbf{Conditional algorithmic information}, denoted $K(x \mid y)$, is the length of the shortest program that generates the string $x$ when given $y$ as input. Intuitively, conditional algorithmic information represents how compressible a string $x$ is given another string $y$. To see how this can work, consider a third string:
\begin{exe}
    \ex\label{ex:bitflip} $c=0011110001010110011011010...$
\end{exe}
\noindent This string is also random, so the algorithmic information $K(c)$ is large. However, $c$ is the result of flipping each bit in $b$, and therefore if $b$ is provided as input, then there \emph{is} a simple program that will yield the string $c$ as output, and thus the program length $K(c \mid b)$ will be short.

With those two definitions we can now define the \textbf{algorithmic mutual information} as the difference between the unconditional and conditional algorithmic information:
\begin{equation}
\label{eq:mai}
    I_K(y; x) = K(x) - K(x|y).
\end{equation}
Algorithmic mutual information measures shared structure between $x$ and $y$. When $K(x)$ is large and $K(x \mid y)$ is small, then it can be said that the two strings share a lot of structural information, because having access to $y$ enables compression of $x$. However, if $K(x)$ and $K(x \mid y)$ are similar or equal in size then we can say that $y$ shares very little structural information with $x$, because knowing $x$ does not contribute to making $y$ more compressible. Note that unlike Shannon mutual information \citep{cover2006elements}, algorithmic mutual information is not commutative: in general, $I_K(x;y) \neq I_K(y;x)$.

We will not apply algorithmic mutual information directly in our measurements and comparisons of orthographic transparency. This reason is that it is difficult to use algorithmic mutual information when dealing with datasets of varying complexities. A low $K(x)$ places a low ceiling on mutual algorithmic information. In our study this would lead to unintuitive interactions between what we would like to measure---orthographic transparency---and extraneous factors including phonotactic and morphological complexity. For this reason we suggest a normalized measure which we dub \textbf{mutual compressibility}, or $C_K(y; x)$. We define mutual compressibility as the \emph{proportion} of $x$ explained by $y$, or

\begin{equation}
    \label{eq:mc}
    C_K(y; x) = \frac{I_K(y; x)}{K(x)} = 1 - \frac{K(x \mid y)}{K(x)} .
\end{equation}

In order to apply these ideas to orthographic transparency, we will consider how much mutual compressibility exists between a language's orthography and its phonology. The motivation here is simple: we consider orthographic transparency as the complexity of the decoding process that converts phonology to orthography, or vice versa. Simple decoding processes are simple programs, but they also imply a shared structure in the data, a fact we will return to later.

\begin{figure}[t!]
    \centering
    \includegraphics[width=1\linewidth]{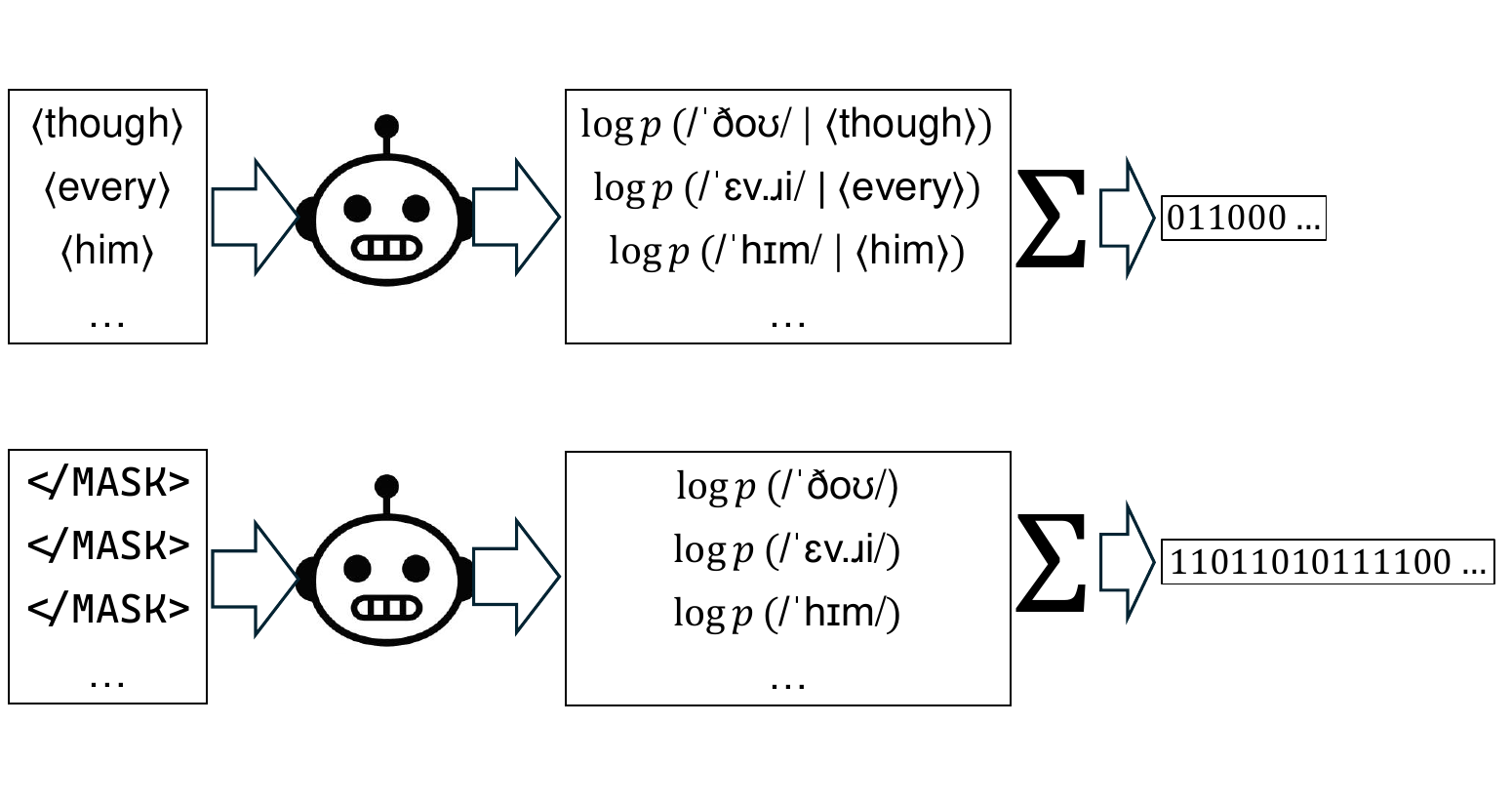}
    \caption{Phonological datasets are compressed both with \textbf{Top}: orthographic side-information and \textbf{Bottom:} no additional information. The difference between the lengths in the two compressed datasets yields the mutual algorithmic information between the dataset and the side information. We show this to be a good approximation of orthographic transparency.}
    \label{fig:method}
\end{figure}

\subsection{Prequential Coding}

\begin{figure}[t!]
    \centering
    \includegraphics[width=\columnwidth]{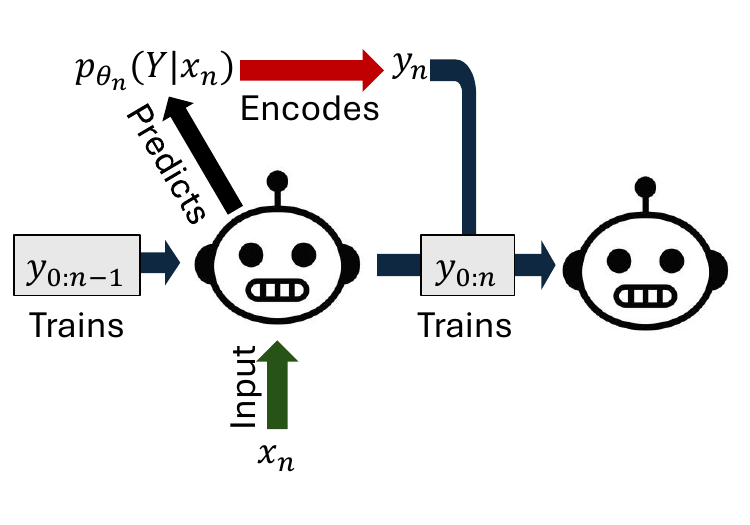}
    \caption{A depiction of prequential coding. Labels and inputs from previous samples at $t = 1, 2,  ..., n-1$ are used to train the current model and yield parameterization $\theta_n$. The current model is used to make a prediction given input $x_n$. The resulting distribution is used to encode label $y_n$.}
    \label{fig:prequential-coding}
    \includegraphics[width=\columnwidth]{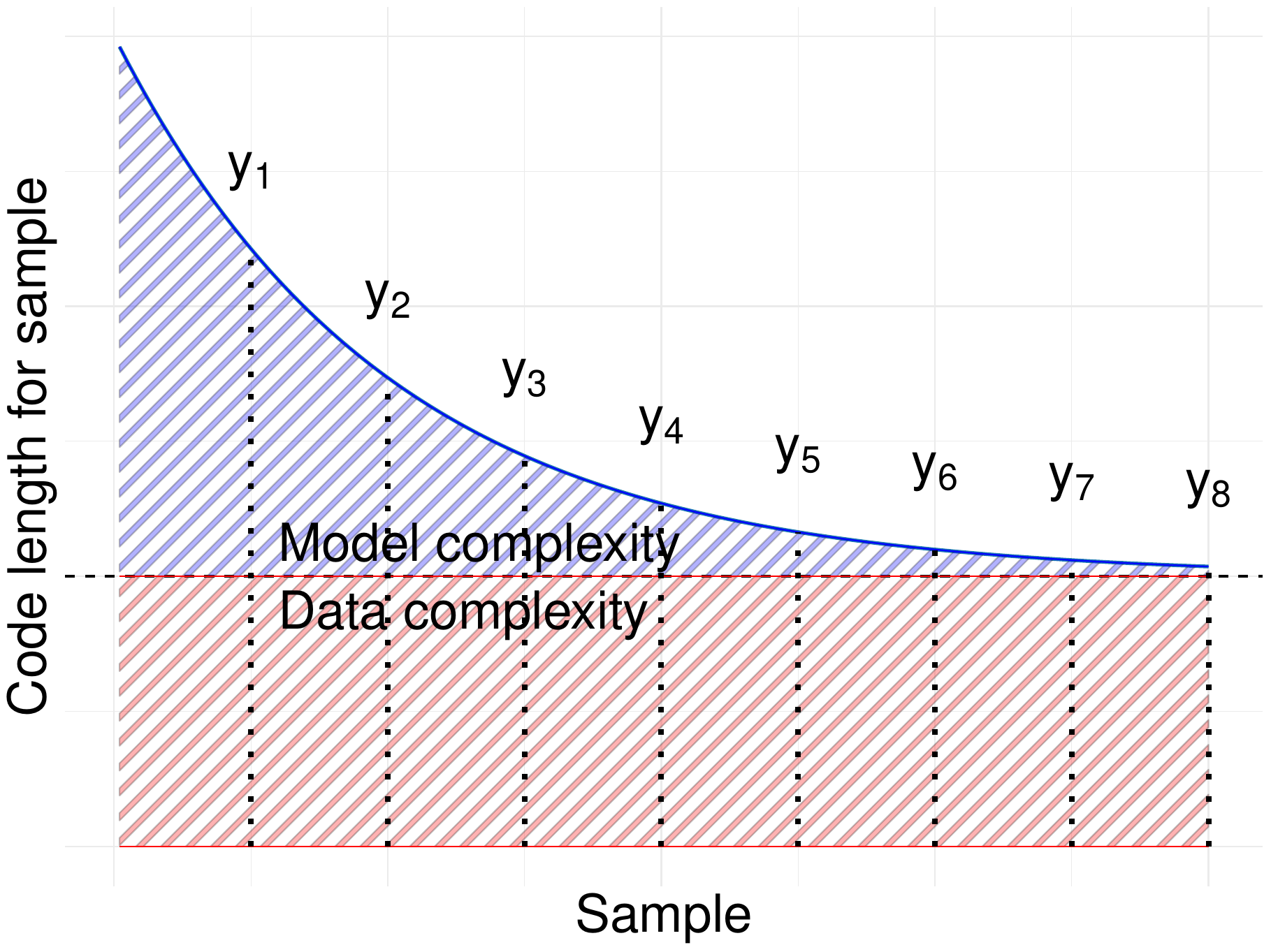}
    \caption{When predicting successive labels a model improves with successive samples. The prequential code is the sum of codelengths for labels $y_n$. This codelength includes both the complexity of the learning task and the inherent entropy of the encoded data. For our specific needs, the complexity of the learning task corresponds to opacity from rule complexity, and the complexity of the dataset reflects irregularity.}
    \label{fig:prequential-curve}
\end{figure}

On their own, the algorithmic complexity measures above are uncomputable in general, because finding such a measurement would require searching through all halting computer programs on a Turing machine to find the shortest program that generates a certain output \citep{li2008introduction}. That being the case, we must rely on approximations of these functions. We will use \textbf{prequential coding}.

Prequential coding, discovered by \citet{dawid1984present} and \citet{rissanen1998stochastic} independently, is a method of approximating the algorithmic complexity of a dataset using a statistical model. 
To demonstrate how prequential coding works, imagine that you would like to communicate an unknown dataset to a partner (this dataset could be a sequence, or a set of labels for some shared inputs). Assuming that you and your partner can agree on a model class to use to encode the data, what is the optimal code for that data? If the data is known ahead of time, then the optimal code would be the maximum likelihood estimate for the model class selected given the dataset. However, without knowing the data ahead of time this cannot work, so instead we choose the next best thing: an incremental code which is determined by the maximum likelihood parameterization \textit{given by the data seen so far}, as illustrated in Figure~\ref{fig:prequential-coding}. This yields the following code length:
\begin{equation}
\label{eq:prequentialcode}
    L_{p_\theta}(x) = -\sum_{i=1}^{N} \log p(x_{i} \mid \theta(x_{1:i-1})),
\end{equation}
where $N$ is the length of the dataset, $p$ is the family of models chosen for the task, and $\theta(x_{1:i-1})$ is the maximum likelihood estimate for the model parameters $\theta$ given observations up to but not including the $i$'th. The logic is illustrated in Figure~\ref{fig:prequential-curve}.


Prequential coding provides a means of estimating the algorithmic information $K(x)$ given a statistical model class, and it has already been used in the literature for this purpose. However, we aim to use it in the estimation of algorithmic mutual information and mutual compressibility. This involves the calculation of two terms---$L_{p_\theta}(x)$ and $L_{p_\theta}(x \mid y)$---which we slot into Equation~\ref{eq:mai}. That is, the estimate of mutual compressibility for output string $x$ given input string $y$ using prequential coding is
\begin{equation}
\label{eq:prequentialcomp}
C_{p_\theta}(y; x) = 1 - \frac{L_{p_\theta}(x|y)}{L_{p_\theta}(x)}.
\end{equation}

In all of our experiments below, $p_\theta$ will be a neural network parameterized by $\theta$. 
We use the Mini-batch Incremental/Replay Streams approach to estimating the prequential code implemented in \citet{bornschein2022sequential}, as standard prequential coding would involve training to convergence on every batch, a procedure which scales quadratically with the length of the data.

In addition to providing a convenient method to estimate mutual compressibility, prequential coding also demonstrates how algorithmic mutual information reflects the two sources of opacity discussed in Section~\ref{sec:background}: irregularity of spellings and complexity of rules. In particular, the `data complexity' component in Figure~\ref{fig:prequential-curve} corresponds to irregularity, and the `model complexity' component corresponds to rule complexity. For example, even if one has fully mastered all the regular rules of English spelling, the one-off irregular pronunciation of \texttt{colonel} as /\textipa{k@\textrhoticity n@l}/ will still be surprising if this word has never been seen before, contributing to data complexity. Conversely, while French spelling is quite regular, its rules are complex, meaning that a learner will take many samples to figure out the rules, contributing to model complexity. Thus, we see that the mutual compressibility measure provides a principled way to combine sources of opacity into a single notion of complexity.


\section{Methods and Dataset}

Here we describe our crosslinguistic dataset of paired orthography and phonetic transcriptions, and our methods for measuring orthographic transparency as mutual compressibility of orthography and phonology.

\subsection{Dataset Source}

We use WikiPron \citep{lee2020wikipron},\footnote{\url{https://github.com/CUNY-CL/wikipron}} a dataset of orthographic forms paired with IPA transcriptions scraped from Wiktionary,\footnote{Available under a CC-BY-SA license.} a crowdsourced online dictionary. For Japanese phonetic information we supplemented WikiPron with an online IPA dictionary\footnote{\url{https://github.com/open-dict-data/ipa-dict}} which took its Japanese data from \citet{breen2013edict}. This was done to provide broad transcriptions since Japanese was of particular interest to us. We reduce the WikiPron corpus for each language to the same sample size by taking the $N=5032$ most frequent forms per language,\footnote{$5032$ was the smallest number of types found in any dataset with a number of types $\geq5000$, our chosen cutoff.} with frequency ranks derived from the most recent 300k sentences collection from the Leipzig news corpus \citep{goldhahn2012building}.\footnote{\url{https://wortschatz.uni-leipzig.de/en/download}, available under a CC-BY license.} The reason for taking the most frequent forms is that highly infrequent forms might be nonrepresentative in terms of orthography and phonotactics. Additionally, orthographic representations with multiple phonetic representations in Wikipron are reduced, and the first phonetic representation in the set was the only one preserved. Although this causes a loss of an important source of irregularity in our measurement (for example, a word like \texttt{lead} being restricted to only one pronunciation in our sets), it eliminates a major source of spurious irregularity (the representation of multiple dialectical pronunciations in the data).

\subsection{Broad and narrow transcriptions}

When assembling our datasets we used broad transcriptions where possible. Where this was not possible we used the narrow transcriptions provided. We made one exception for Japanese, as mentioned earlier, where a different transcription was sourced in order to use a broader transcription type than Wiktionary could provide. Breadth of transcription is a source of additional---and difficult to control---complexity, so we did our best to keep them consistent across languages to the extent our data source allowed.

WikiPron, being a scrape of Wiktionary, contains many different crowdsourced phonetic transcriptions. Though it serves as an excellent source when it comes to breadth, there can be high variance in the specificity of the transcriptions. As an example, consider the following two transcriptions for the same English word \texttt{winter} (both found in the WikiPron set).

\begin{exe}
    
    \ex /\textipa{"wInt\textrhookschwa}/
    \label{broad}
    
    \ex \textipa{["wI\~R\textrhookschwa]}
    \label{narrow}
\end{exe}

\noindent Example~\ref{broad} transcribes \texttt{winter} broadly, without representing additional phonological processes that may be operating when pronouncing the word. Example~\ref{narrow} on the other hand is written such that it includes symbols which denote movements a speaker makes in producing this word which may be phonologically conditioned and predictable. For example, the narrow transcription uses the symbol \textipa{R} to denote the rapid flapping motion an English speaker will often use to produce a /t/ in that particular phonetic environment. While technically a spectrum (as one can transcribe movements with increasing levels of detail) Wiktionary makes a two-way distinction between these trascription types reflecting the broader tendency in linguistics to denote phonemic and phonological levels of detail.

\subsection{Treatment of sub-character structure}

We preprocessed our dataset to decompose characters where they contained subcomponents that would normally be accessible to a human reader. To this end we performed decompositions on Japanese and Chinese characters \citep{kishore2016makemeahanzi},\footnote{\url{https://github.com/skishore/makemeahanzi}} jamo decompositions on Korean Hangul \citep{dong2018pythonjamo},\footnote{\url{https://github.com/JDongian/python-jamo}} and NFD normalization on all characters \citep{Unicode15}. To further explain why, consider first Chinese, Japanese, and Korean. Chinese characters, kanji, and Hangul all have compositional structure which is not reflected in their UTF-8 encodings \citep{Unicode15}. Many Chinese characters (and kanji) contain phonetic elements which are partially informative about pronunciation in Chinese. The case of Hangul is even more important, as Hangul are composed of component jamo, which are not usually represented in Unicode formatted text (including our datasets), yet are meant to represent particular phonemes. 
These processing techniques are our best attempts at preserving the internal structure of characters that is accessible to humans, in order to get a human-applicable measure from neural networks. 

\subsection{Implementation of Prequential Coding}

We now turn to the estimation of mutual compressibility in the dataset using prequential coding. We use a sequence-to-sequence CNN as the underlying statistical model \citep{gehring2017convolutional}. The CNN has a single encoding and decoding layer, and all hidden layers are sized to $100$ dimensions. 

The actual prequential coding procedure we use---Mini-batch Incremental with Replay Streams \citep[MI/RS]{bornschein2022sequential}---is an approximate way to find the true prequential codelength. True prequential coding requires training to convergence on all prefixes of the data to obtain maximum likelihood estimates \citep{rissanen1998stochastic}; however, the procedure's time complexity scales quadratically in the length of the dataset to compress. MI/RS, on the other hand, has a time complexity of $O(kn)$. This approximation works by keeping a series of $k$ replay streams running on previously seen data with a probability $p$ of resetting each stream after each training iteration to iteration 1. Details of the implementation can be found in \citet{bornschein2022sequential}, but for all of our coding procedures we run $k=25$ replay streams with a probability $p=\frac{1}{i+1}$ of resetting at each iteration $i$ (a setting which ensures uniform sampling over all data as $i$ increases when indexed at 1).

To obtain the compressibility measurements, MI/RS must be run once with side information and once without. What this means in practice is running the sequence-to-sequence CNN with and without input information masked. This procedure yields the two terms in Equation~\ref{eq:prequentialcomp}. When we perform both the masked and unmasked runs, we initialize the models with the same seeds both times. We repeat this procedure 40 times (with 40 different seeds) for each language, and for each direction (phonology-to-orthography and orthography-to-phonology). We never really compress the datasets, but the distributions predicted by our models allow us to calculate the length (in bits) of the ideal encoding given by those distributions via Equation~\ref{eq:prequentialcode}.

\section{Results}

\subsection{Main Results}

We present average mutual compressibility for orthography from phonology, and vice versa, for 22 languages in our WikiPron dataset. 

\begin{figure*}[h]
    \centering
    \includegraphics[width=\textwidth]{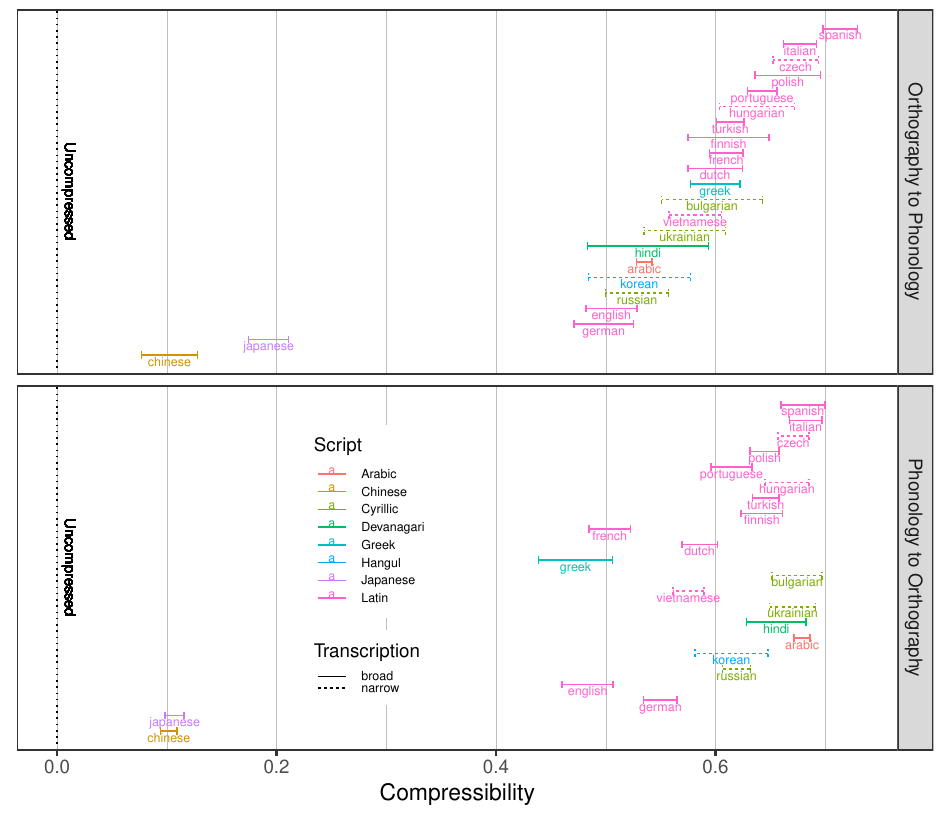}
    \caption{Confidence intervals for mutual compressibility of orthography given phonology, and vice versa, in 22 languages. Error bars represent 95\% confidence intervals as determined over all $40$ runs.}
    \label{fig:main}
\end{figure*}

Figure~\ref{fig:main} collects main results. At a high level, we observe some encouraging patterns. For example, as a sanity check, we should expect phonological systems to be more compressible than logographic ones. This is the case, with Chinese and Japanese both scoring extremely low in the compressibility metric.

We can also observe cross-directional effects with some languages showing significant differences between the orthography-to-phonology mapping and the phonology-to-orthography mapping. Although some past work has made the claim that spelling (that is, phonology-to-orthography) is more difficult than reading in general \citep{bosman1997spelling}, we can see this doesn't always hold in terms of mutual compressibility. While some languages like French or Greek exhibit the expected effect, there are many languages which exhibit the reverse phenomenon. In many cases this appears to be due to the transcription form in the dataset---with narrow transcriptions making the ``reading" mapping more difficult---thus making the results a likely artifact of the data. But this is not always the case. For example, German exhibits this phenomenon and its transcription is broad. More interestingly, for some languages this directional difficulty may be inherent to the writing system. Hindi and Arabic both show this effect, and both have at least occasional optional vowel writing, with Hindi having implicit vowels when unmarked \citep{snell2014complete}, and with Arabic leaving short vowels unwritten in most contexts. We will go review more specifics as when we compare our results to the empirical data. We provide a table comparing the languages in Appendix~\ref{appendix:sig} for those interested in more detail.

\subsection{The Case of Japanese}

\begin{figure}[h!]
    \centering
    \includegraphics[width=\columnwidth]{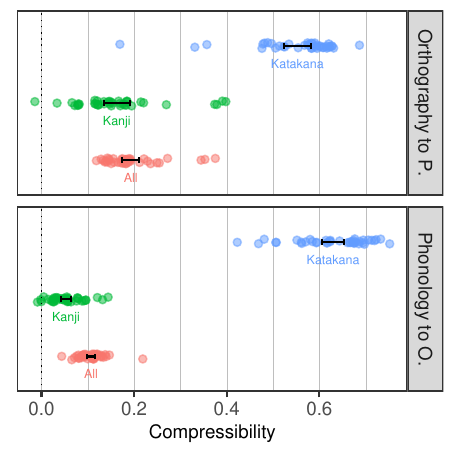}
    \caption{Within languages the measure also can provide distinctions between script types. Here we compare katakana, kanji, and the full language including all three scripts. Hiragana was excluded because there were not enough stand-alone words made exclusively of hiragana characters to analyze.}
    \label{fig:japanese}
\end{figure}

We want to highlight the case of Japanese, in particular, to show that our approach delivers sensible results. With our method we can both get a measure of the unified transparency of the whole system, and by filtering by character type, a measure of the transparency of individual scripts. This makes Japanese an interesting case. The system is a hybrid of three scripts: kanji, katakana, and hiragana. Two scripts, katakana and hiragana, are syllabic scripts, meaning single characters stand for whole syllables. One script, kanji, is composed of Chinese character borrowings and is thus logographic. What is the transparency of individual component scripts, and how does it affect the overall unified system? To investigate this question we analyzed katakana, kanji, and the union of all systems.\footnote{Hiragana was dropped as an individual system because not enough stand-alone hiragana-only words were present in our dataset matching our $N\geq5000$ cutoff.} 

The plots of this split can be seen in Figure~\ref{fig:japanese}.
As one would hope, the syllabic script katakana shows a much higher transparency in both directions. Kanji are less compressible, with the overall orthography intermediate between the two. This is not too surprising, but interestingly we can also see directional effects. Katakana is relatively transparent in both directions, with an effect perhaps favoring the phonology-to-orthography direction. Kanji, however, shows ease in the orthography-to-phonology direction. We hypothesize that some kanji may have transparency if pronounced with the Chinese readings. Since our method only preserved a single pronunciation per kanji, it is possible that this effect could be exaggerated by our preprocessing.

\subsection{Comparison with previous measures}

\begin{figure}
    \centering
    \includegraphics[width=\columnwidth]{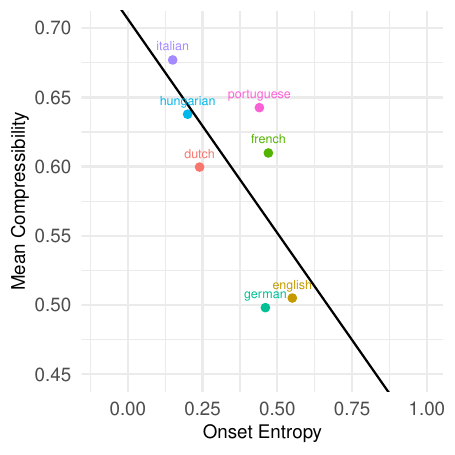}
    \caption{Our measure plotted against the onset entropy values reported in \citet{borgwaldt2005onset}. Onset entropy is the unweighted mean of the Shannon entropy of the first phoneme in a word given the first grapheme.}
    \label{fig:empirical}
\end{figure}

For languages for which there is existing data, the results largely conform to existing beliefs about how these languages would rank in terms of transparency.  For example, there is relative agreement about the high transparency of Italian, Finnish, and Hungarian \citep{seymour2003foundation}. There is also agreement that English is relatively opaque among the alphabetic systems, and among these English does rank lowest in our metric. We have also reproduced certain directional effects. For example, the strong directionality effects in Greek reported by \citet{protopapas2009greek} can be seen. The same goes for the directional effects observed in French as in \citep{ziegler1996statistical}. Against onset entropy \citep{borgwaldt2005onset}, our measure seems to perform similarly as plotted in Figure~\ref{fig:empirical}, showing high compressibility when there is low entropy as in Italian and Hungarian, and vice versa as in English, though the relationship is not significant given the small number of datapoints.

We note however that this is not always the case. Korean, often considered quite transparent, ranks as middling-to-low among the phonologically based writing systems in Figure~\ref{fig:main}. This is plausibly an issue with the dataset, as the phonological dataset for Korean is extremely narrowly transcribed. What this means is that the model, while learning the Korean mappings, mustn't just recover the grapheme-to-phoneme mappings, but the phoneme-to-allophone mappings as well. Equally surprising to us was the case of German, as in the literature it is often contrasted with English as relatively transparent, yet in the orthography-to-phonology direction equally opaque. We suspect this reflects loanwords which retain the source language orthography.

\section{Conclusion}

We have demonstrated that mutual compressibility provides a sensible measure of orthographic transparency that is more general than previous proposals, conferring the ability to analyze multiple kinds of systems (logographic, syllabic, alphabetic, and others) via a unified metric. We also believe that this approach is strong on theoretical grounds, based on the intuition that transparency is related to shared structure between different modalities. Finally, we believe that it has some explanatory power. Difficult-to-compress concepts are difficult to learn, a known bias which has been demonstrated in different domains \citep{feldman2003simplicity, chater2003simplicity} and would explain the difficulties encountered in learning to read opaque orthographies \citep{seymour2003foundation, aro2003learning}.

\section*{Limitations}
The present method has some practical limitations. Firstly, the Wiktionary data varies by transcription narrowness. This makes comparisons between languages less accurate than they might otherwise be, and they could be theoretically improved by adopting transcriptions from a more consistent source than a Wiktionary scrape. There are also more mundane engineering questions. There is the model selection question: Can we home in on more accurate scores---and more humanlike ones---by using a better class of statistical model? There is also the question of the MI/RS method for prequential estimation. We could use other methods (like block encoding, \citealt{blier2018description}) or optimize the hyperparameters to achieve tighter bounds on the true measure. We believe these issues worthy of future investigations.

One potential confounding source of opacity could be in the genre of writing itself. We use news corpora to provide frequency information. News sources, especially in what might be termed exoteric communities \citep{wray2007consequences}, can incorporate many loan words and foreign names, whose presence in the data could impact opacity scores. A more careful analysis of genre-related effects, and of effects from particular sources of words, is also warranted. 


\section*{Ethical Considerations} We foresee no ethical issues with this work.

\bibliography{custom}

\appendix

\section{Between-Language Significance Tests}
\label{appendix:sig}
We report in Table~\ref{tab:tukey-full} the results of TukeyHSD tests on both the orthography-to-phonology and phonology-to-orthography mappings for those interested. Colored cells are significant with $p<.05$.

\begin{table*}[t!]
\centering
\begin{minipage}[t!]{0.82\textwidth}
\begin{subtable}[t!]{\linewidth}
\centering
\caption{Orthography differences}
\label{tab:tukeyorth}
\resizebox{\linewidth}{!}{
\begin{tabular}{lcccccccccccccccccccccc}
\toprule
 & ara & bul & ces & deu & ell & eng & fin & fra & hin & hun & ita & jpn & kor & nld & pol & por & rus & spa & tur & ukr & vie & zho
\\
\midrule
ara & --- & \heatcell{-0.062}{0.291} & \heatcell{-0.138}{0.000} & \heatcell{0.037}{0.979} & \heatcell{-0.065}{0.216} & \heatcell{0.030}{0.998} & \heatcell{-0.077}{0.039} & \heatcell{-0.075}{0.055} & \heatcell{-0.003}{1.000} & \heatcell{-0.103}{0.000} & \heatcell{-0.142}{0.000} & \heatcell{0.342}{0.000} & \heatcell{0.004}{1.000} & \heatcell{-0.065}{0.216} & \heatcell{-0.131}{0.000} & \heatcell{-0.107}{0.000} & \heatcell{0.007}{1.000} & \heatcell{-0.179}{0.000} & \heatcell{-0.078}{0.030} & \heatcell{-0.037}{0.980} & \heatcell{-0.046}{0.826} & \heatcell{0.433}{0.000}
\\
bul & \heatcell{0.062}{0.291} & --- & \heatcell{-0.076}{0.044} & \heatcell{0.099}{0.001} & \heatcell{-0.003}{1.000} & \heatcell{0.092}{0.002} & \heatcell{-0.015}{1.000} & \heatcell{-0.013}{1.000} & \heatcell{0.058}{0.402} & \heatcell{-0.041}{0.940} & \heatcell{-0.080}{0.023} & \heatcell{0.404}{0.000} & \heatcell{0.066}{0.176} & \heatcell{-0.003}{1.000} & \heatcell{-0.069}{0.123} & \heatcell{-0.046}{0.845} & \heatcell{0.068}{0.135} & \heatcell{-0.117}{0.000} & \heatcell{-0.017}{1.000} & \heatcell{0.025}{1.000} & \heatcell{0.015}{1.000} & \heatcell{0.495}{0.000}
\\
ces & \heatcell{0.138}{0.000} & \heatcell{0.076}{0.044} & --- & \heatcell{0.175}{0.000} & \heatcell{0.073}{0.067} & \heatcell{0.168}{0.000} & \heatcell{0.061}{0.315} & \heatcell{0.063}{0.252} & \heatcell{0.134}{0.000} & \heatcell{0.035}{0.988} & \heatcell{-0.004}{1.000} & \heatcell{0.480}{0.000} & \heatcell{0.142}{0.000} & \heatcell{0.073}{0.067} & \heatcell{0.007}{1.000} & \heatcell{0.030}{0.998} & \heatcell{0.144}{0.000} & \heatcell{-0.041}{0.942} & \heatcell{0.059}{0.366} & \heatcell{0.101}{0.000} & \heatcell{0.091}{0.003} & \heatcell{0.571}{0.000}
\\
deu & \heatcell{-0.037}{0.979} & \heatcell{-0.099}{0.001} & \heatcell{-0.175}{0.000} & --- & \heatcell{-0.102}{0.000} & \heatcell{-0.007}{1.000} & \heatcell{-0.114}{0.000} & \heatcell{-0.112}{0.000} & \heatcell{-0.040}{0.947} & \heatcell{-0.140}{0.000} & \heatcell{-0.179}{0.000} & \heatcell{0.305}{0.000} & \heatcell{-0.033}{0.996} & \heatcell{-0.102}{0.000} & \heatcell{-0.168}{0.000} & \heatcell{-0.144}{0.000} & \heatcell{-0.030}{0.998} & \heatcell{-0.216}{0.000} & \heatcell{-0.115}{0.000} & \heatcell{-0.074}{0.062} & \heatcell{-0.083}{0.013} & \heatcell{0.396}{0.000}
\\
ell & \heatcell{0.065}{0.216} & \heatcell{0.003}{1.000} & \heatcell{-0.073}{0.067} & \heatcell{0.102}{0.000} & --- & \heatcell{0.095}{0.001} & \heatcell{-0.012}{1.000} & \heatcell{-0.010}{1.000} & \heatcell{0.061}{0.311} & \heatcell{-0.038}{0.970} & \heatcell{-0.077}{0.036} & \heatcell{0.407}{0.000} & \heatcell{0.069}{0.124} & \heatcell{-0.000}{1.000} & \heatcell{-0.066}{0.174} & \heatcell{-0.043}{0.907} & \heatcell{0.071}{0.093} & \heatcell{-0.114}{0.000} & \heatcell{-0.014}{1.000} & \heatcell{0.028}{1.000} & \heatcell{0.018}{1.000} & \heatcell{0.497}{0.000}
\\
eng & \heatcell{-0.030}{0.998} & \heatcell{-0.092}{0.002} & \heatcell{-0.168}{0.000} & \heatcell{0.007}{1.000} & \heatcell{-0.095}{0.001} & --- & \heatcell{-0.107}{0.000} & \heatcell{-0.105}{0.000} & \heatcell{-0.033}{0.994} & \heatcell{-0.133}{0.000} & \heatcell{-0.172}{0.000} & \heatcell{0.312}{0.000} & \heatcell{-0.026}{1.000} & \heatcell{-0.095}{0.001} & \heatcell{-0.161}{0.000} & \heatcell{-0.138}{0.000} & \heatcell{-0.023}{1.000} & \heatcell{-0.209}{0.000} & \heatcell{-0.108}{0.000} & \heatcell{-0.067}{0.163} & \heatcell{-0.076}{0.042} & \heatcell{0.403}{0.000}
\\
fin & \heatcell{0.077}{0.039} & \heatcell{0.015}{1.000} & \heatcell{-0.061}{0.315} & \heatcell{0.114}{0.000} & \heatcell{0.012}{1.000} & \heatcell{0.107}{0.000} & --- & \heatcell{0.002}{1.000} & \heatcell{0.073}{0.066} & \heatcell{-0.026}{1.000} & \heatcell{-0.065}{0.204} & \heatcell{0.419}{0.000} & \heatcell{0.081}{0.018} & \heatcell{0.012}{1.000} & \heatcell{-0.054}{0.565} & \heatcell{-0.031}{0.998} & \heatcell{0.083}{0.013} & \heatcell{-0.102}{0.000} & \heatcell{-0.002}{1.000} & \heatcell{0.040}{0.952} & \heatcell{0.030}{0.998} & \heatcell{0.510}{0.000}
\\
fra & \heatcell{0.075}{0.055} & \heatcell{0.013}{1.000} & \heatcell{-0.063}{0.252} & \heatcell{0.112}{0.000} & \heatcell{0.010}{1.000} & \heatcell{0.105}{0.000} & \heatcell{-0.002}{1.000} & --- & \heatcell{0.071}{0.090} & \heatcell{-0.028}{0.999} & \heatcell{-0.067}{0.157} & \heatcell{0.417}{0.000} & \heatcell{0.079}{0.027} & \heatcell{0.010}{1.000} & \heatcell{-0.056}{0.483} & \heatcell{-0.033}{0.995} & \heatcell{0.081}{0.019} & \heatcell{-0.104}{0.000} & \heatcell{-0.004}{1.000} & \heatcell{0.038}{0.973} & \heatcell{0.028}{0.999} & \heatcell{0.507}{0.000}
\\
hin & \heatcell{0.003}{1.000} & \heatcell{-0.058}{0.402} & \heatcell{-0.134}{0.000} & \heatcell{0.040}{0.947} & \heatcell{-0.061}{0.311} & \heatcell{0.033}{0.994} & \heatcell{-0.073}{0.066} & \heatcell{-0.071}{0.090} & --- & \heatcell{-0.099}{0.000} & \heatcell{-0.139}{0.000} & \heatcell{0.346}{0.000} & \heatcell{0.008}{1.000} & \heatcell{-0.061}{0.310} & \heatcell{-0.128}{0.000} & \heatcell{-0.104}{0.000} & \heatcell{0.010}{1.000} & \heatcell{-0.175}{0.000} & \heatcell{-0.075}{0.052} & \heatcell{-0.033}{0.994} & \heatcell{-0.043}{0.905} & \heatcell{0.436}{0.000}
\\
hun & \heatcell{0.103}{0.000} & \heatcell{0.041}{0.940} & \heatcell{-0.035}{0.988} & \heatcell{0.140}{0.000} & \heatcell{0.038}{0.970} & \heatcell{0.133}{0.000} & \heatcell{0.026}{1.000} & \heatcell{0.028}{0.999} & \heatcell{0.099}{0.000} & --- & \heatcell{-0.039}{0.961} & \heatcell{0.445}{0.000} & \heatcell{0.107}{0.000} & \heatcell{0.038}{0.970} & \heatcell{-0.028}{0.999} & \heatcell{-0.005}{1.000} & \heatcell{0.109}{0.000} & \heatcell{-0.076}{0.045} & \heatcell{0.024}{1.000} & \heatcell{0.066}{0.184} & \heatcell{0.056}{0.478} & \heatcell{0.535}{0.000}
\\
ita & \heatcell{0.142}{0.000} & \heatcell{0.080}{0.023} & \heatcell{0.004}{1.000} & \heatcell{0.179}{0.000} & \heatcell{0.077}{0.036} & \heatcell{0.172}{0.000} & \heatcell{0.065}{0.204} & \heatcell{0.067}{0.157} & \heatcell{0.139}{0.000} & \heatcell{0.039}{0.961} & --- & \heatcell{0.484}{0.000} & \heatcell{0.146}{0.000} & \heatcell{0.077}{0.036} & \heatcell{0.011}{1.000} & \heatcell{0.034}{0.991} & \heatcell{0.148}{0.000} & \heatcell{-0.037}{0.980} & \heatcell{0.063}{0.243} & \heatcell{0.105}{0.000} & \heatcell{0.095}{0.001} & \heatcell{0.575}{0.000}
\\
jpn & \heatcell{-0.342}{0.000} & \heatcell{-0.404}{0.000} & \heatcell{-0.480}{0.000} & \heatcell{-0.305}{0.000} & \heatcell{-0.407}{0.000} & \heatcell{-0.312}{0.000} & \heatcell{-0.419}{0.000} & \heatcell{-0.417}{0.000} & \heatcell{-0.346}{0.000} & \heatcell{-0.445}{0.000} & \heatcell{-0.484}{0.000} & --- & \heatcell{-0.338}{0.000} & \heatcell{-0.407}{0.000} & \heatcell{-0.473}{0.000} & \heatcell{-0.450}{0.000} & \heatcell{-0.336}{0.000} & \heatcell{-0.521}{0.000} & \heatcell{-0.421}{0.000} & \heatcell{-0.379}{0.000} & \heatcell{-0.389}{0.000} & \heatcell{0.090}{0.003}
\\
kor & \heatcell{-0.004}{1.000} & \heatcell{-0.066}{0.176} & \heatcell{-0.142}{0.000} & \heatcell{0.033}{0.996} & \heatcell{-0.069}{0.124} & \heatcell{0.026}{1.000} & \heatcell{-0.081}{0.018} & \heatcell{-0.079}{0.027} & \heatcell{-0.008}{1.000} & \heatcell{-0.107}{0.000} & \heatcell{-0.146}{0.000} & \heatcell{0.338}{0.000} & --- & \heatcell{-0.069}{0.124} & \heatcell{-0.135}{0.000} & \heatcell{-0.112}{0.000} & \heatcell{0.002}{1.000} & \heatcell{-0.183}{0.000} & \heatcell{-0.083}{0.014} & \heatcell{-0.041}{0.935} & \heatcell{-0.051}{0.682} & \heatcell{0.428}{0.000}
\\
nld & \heatcell{0.065}{0.216} & \heatcell{0.003}{1.000} & \heatcell{-0.073}{0.067} & \heatcell{0.102}{0.000} & \heatcell{0.000}{1.000} & \heatcell{0.095}{0.001} & \heatcell{-0.012}{1.000} & \heatcell{-0.010}{1.000} & \heatcell{0.061}{0.310} & \heatcell{-0.038}{0.970} & \heatcell{-0.077}{0.036} & \heatcell{0.407}{0.000} & \heatcell{0.069}{0.124} & --- & \heatcell{-0.066}{0.174} & \heatcell{-0.043}{0.907} & \heatcell{0.071}{0.093} & \heatcell{-0.114}{0.000} & \heatcell{-0.014}{1.000} & \heatcell{0.028}{1.000} & \heatcell{0.018}{1.000} & \heatcell{0.497}{0.000}
\\
pol & \heatcell{0.131}{0.000} & \heatcell{0.069}{0.123} & \heatcell{-0.007}{1.000} & \heatcell{0.168}{0.000} & \heatcell{0.066}{0.174} & \heatcell{0.161}{0.000} & \heatcell{0.054}{0.565} & \heatcell{0.056}{0.483} & \heatcell{0.128}{0.000} & \heatcell{0.028}{0.999} & \heatcell{-0.011}{1.000} & \heatcell{0.473}{0.000} & \heatcell{0.135}{0.000} & \heatcell{0.066}{0.174} & --- & \heatcell{0.023}{1.000} & \heatcell{0.137}{0.000} & \heatcell{-0.048}{0.786} & \heatcell{0.052}{0.623} & \heatcell{0.094}{0.002} & \heatcell{0.084}{0.010} & \heatcell{0.564}{0.000}
\\
por & \heatcell{0.107}{0.000} & \heatcell{0.046}{0.845} & \heatcell{-0.030}{0.998} & \heatcell{0.144}{0.000} & \heatcell{0.043}{0.907} & \heatcell{0.138}{0.000} & \heatcell{0.031}{0.998} & \heatcell{0.033}{0.995} & \heatcell{0.104}{0.000} & \heatcell{0.005}{1.000} & \heatcell{-0.034}{0.991} & \heatcell{0.450}{0.000} & \heatcell{0.112}{0.000} & \heatcell{0.043}{0.907} & \heatcell{-0.023}{1.000} & --- & \heatcell{0.114}{0.000} & \heatcell{-0.071}{0.092} & \heatcell{0.029}{0.999} & \heatcell{0.071}{0.099} & \heatcell{0.061}{0.313} & \heatcell{0.540}{0.000}
\\
rus & \heatcell{-0.007}{1.000} & \heatcell{-0.068}{0.135} & \heatcell{-0.144}{0.000} & \heatcell{0.030}{0.998} & \heatcell{-0.071}{0.093} & \heatcell{0.023}{1.000} & \heatcell{-0.083}{0.013} & \heatcell{-0.081}{0.019} & \heatcell{-0.010}{1.000} & \heatcell{-0.109}{0.000} & \heatcell{-0.148}{0.000} & \heatcell{0.336}{0.000} & \heatcell{-0.002}{1.000} & \heatcell{-0.071}{0.093} & \heatcell{-0.137}{0.000} & \heatcell{-0.114}{0.000} & --- & \heatcell{-0.185}{0.000} & \heatcell{-0.085}{0.010} & \heatcell{-0.043}{0.898} & \heatcell{-0.053}{0.604} & \heatcell{0.426}{0.000}
\\
spa & \heatcell{0.179}{0.000} & \heatcell{0.117}{0.000} & \heatcell{0.041}{0.942} & \heatcell{0.216}{0.000} & \heatcell{0.114}{0.000} & \heatcell{0.209}{0.000} & \heatcell{0.102}{0.000} & \heatcell{0.104}{0.000} & \heatcell{0.175}{0.000} & \heatcell{0.076}{0.045} & \heatcell{0.037}{0.980} & \heatcell{0.521}{0.000} & \heatcell{0.183}{0.000} & \heatcell{0.114}{0.000} & \heatcell{0.048}{0.786} & \heatcell{0.071}{0.092} & \heatcell{0.185}{0.000} & --- & \heatcell{0.100}{0.000} & \heatcell{0.142}{0.000} & \heatcell{0.132}{0.000} & \heatcell{0.611}{0.000}
\\
tur & \heatcell{0.078}{0.030} & \heatcell{0.017}{1.000} & \heatcell{-0.059}{0.366} & \heatcell{0.115}{0.000} & \heatcell{0.014}{1.000} & \heatcell{0.108}{0.000} & \heatcell{0.002}{1.000} & \heatcell{0.004}{1.000} & \heatcell{0.075}{0.052} & \heatcell{-0.024}{1.000} & \heatcell{-0.063}{0.243} & \heatcell{0.421}{0.000} & \heatcell{0.083}{0.014} & \heatcell{0.014}{1.000} & \heatcell{-0.052}{0.623} & \heatcell{-0.029}{0.999} & \heatcell{0.085}{0.010} & \heatcell{-0.100}{0.000} & --- & \heatcell{0.042}{0.930} & \heatcell{0.032}{0.996} & \heatcell{0.511}{0.000}
\\
ukr & \heatcell{0.037}{0.980} & \heatcell{-0.025}{1.000} & \heatcell{-0.101}{0.000} & \heatcell{0.074}{0.062} & \heatcell{-0.028}{1.000} & \heatcell{0.067}{0.163} & \heatcell{-0.040}{0.952} & \heatcell{-0.038}{0.973} & \heatcell{0.033}{0.994} & \heatcell{-0.066}{0.184} & \heatcell{-0.105}{0.000} & \heatcell{0.379}{0.000} & \heatcell{0.041}{0.935} & \heatcell{-0.028}{1.000} & \heatcell{-0.094}{0.002} & \heatcell{-0.071}{0.099} & \heatcell{0.043}{0.898} & \heatcell{-0.142}{0.000} & \heatcell{-0.042}{0.930} & --- & \heatcell{-0.010}{1.000} & \heatcell{0.470}{0.000}
\\
vie & \heatcell{0.046}{0.826} & \heatcell{-0.015}{1.000} & \heatcell{-0.091}{0.003} & \heatcell{0.083}{0.013} & \heatcell{-0.018}{1.000} & \heatcell{0.076}{0.042} & \heatcell{-0.030}{0.998} & \heatcell{-0.028}{0.999} & \heatcell{0.043}{0.905} & \heatcell{-0.056}{0.478} & \heatcell{-0.095}{0.001} & \heatcell{0.389}{0.000} & \heatcell{0.051}{0.682} & \heatcell{-0.018}{1.000} & \heatcell{-0.084}{0.010} & \heatcell{-0.061}{0.313} & \heatcell{0.053}{0.604} & \heatcell{-0.132}{0.000} & \heatcell{-0.032}{0.996} & \heatcell{0.010}{1.000} & --- & \heatcell{0.479}{0.000}
\\
zho & \heatcell{-0.433}{0.000} & \heatcell{-0.495}{0.000} & \heatcell{-0.571}{0.000} & \heatcell{-0.396}{0.000} & \heatcell{-0.497}{0.000} & \heatcell{-0.403}{0.000} & \heatcell{-0.510}{0.000} & \heatcell{-0.507}{0.000} & \heatcell{-0.436}{0.000} & \heatcell{-0.535}{0.000} & \heatcell{-0.575}{0.000} & \heatcell{-0.090}{0.003} & \heatcell{-0.428}{0.000} & \heatcell{-0.497}{0.000} & \heatcell{-0.564}{0.000} & \heatcell{-0.540}{0.000} & \heatcell{-0.426}{0.000} & \heatcell{-0.611}{0.000} & \heatcell{-0.511}{0.000} & \heatcell{-0.470}{0.000} & \heatcell{-0.479}{0.000} & ---
\\
\bottomrule
\end{tabular}
}
\end{subtable}

\bigskip

\begin{subtable}[t!]{\linewidth}
\centering
\caption{Phonology differences}
\resizebox{\linewidth}{!}{

\begin{tabular}{lcccccccccccccccccccccc}
\toprule
 & ara & bul & ces & deu & ell & eng & fin & fra & hin & hun & ita & jpn & kor & nld & pol & por & rus & spa & tur & ukr & vie & zho
\\
\midrule
ara & --- & \heatcell{0.005}{1.000} & \heatcell{0.008}{1.000} & \heatcell{0.129}{0.000} & \heatcell{0.207}{0.000} & \heatcell{0.196}{0.000} & \heatcell{0.037}{0.518} & \heatcell{0.175}{0.000} & \heatcell{0.024}{0.987} & \heatcell{0.014}{1.000} & \heatcell{-0.003}{1.000} & \heatcell{0.572}{0.000} & \heatcell{0.064}{0.001} & \heatcell{0.094}{0.000} & \heatcell{0.034}{0.661} & \heatcell{0.064}{0.001} & \heatcell{0.060}{0.004} & \heatcell{-0.001}{1.000} & \heatcell{0.033}{0.714} & \heatcell{0.009}{1.000} & \heatcell{0.104}{0.000} & \heatcell{0.577}{0.000}
\\
bul & \heatcell{-0.005}{1.000} & --- & \heatcell{0.003}{1.000} & \heatcell{0.125}{0.000} & \heatcell{0.202}{0.000} & \heatcell{0.191}{0.000} & \heatcell{0.032}{0.768} & \heatcell{0.171}{0.000} & \heatcell{0.019}{0.999} & \heatcell{0.009}{1.000} & \heatcell{-0.008}{1.000} & \heatcell{0.567}{0.000} & \heatcell{0.060}{0.004} & \heatcell{0.089}{0.000} & \heatcell{0.030}{0.874} & \heatcell{0.060}{0.004} & \heatcell{0.055}{0.013} & \heatcell{-0.005}{1.000} & \heatcell{0.029}{0.905} & \heatcell{0.004}{1.000} & \heatcell{0.099}{0.000} & \heatcell{0.573}{0.000}
\\
ces & \heatcell{-0.008}{1.000} & \heatcell{-0.003}{1.000} & --- & \heatcell{0.121}{0.000} & \heatcell{0.199}{0.000} & \heatcell{0.187}{0.000} & \heatcell{0.029}{0.903} & \heatcell{0.167}{0.000} & \heatcell{0.016}{1.000} & \heatcell{0.006}{1.000} & \heatcell{-0.011}{1.000} & \heatcell{0.564}{0.000} & \heatcell{0.056}{0.010} & \heatcell{0.085}{0.000} & \heatcell{0.026}{0.960} & \heatcell{0.056}{0.010} & \heatcell{0.052}{0.033} & \heatcell{-0.009}{1.000} & \heatcell{0.025}{0.973} & \heatcell{0.001}{1.000} & \heatcell{0.096}{0.000} & \heatcell{0.569}{0.000}
\\
deu & \heatcell{-0.129}{0.000} & \heatcell{-0.125}{0.000} & \heatcell{-0.121}{0.000} & --- & \heatcell{0.077}{0.000} & \heatcell{0.066}{0.000} & \heatcell{-0.092}{0.000} & \heatcell{0.046}{0.116} & \heatcell{-0.106}{0.000} & \heatcell{-0.115}{0.000} & \heatcell{-0.132}{0.000} & \heatcell{0.443}{0.000} & \heatcell{-0.065}{0.001} & \heatcell{-0.036}{0.576} & \heatcell{-0.095}{0.000} & \heatcell{-0.065}{0.001} & \heatcell{-0.069}{0.000} & \heatcell{-0.130}{0.000} & \heatcell{-0.096}{0.000} & \heatcell{-0.120}{0.000} & \heatcell{-0.026}{0.962} & \heatcell{0.448}{0.000}
\\
ell & \heatcell{-0.207}{0.000} & \heatcell{-0.202}{0.000} & \heatcell{-0.199}{0.000} & \heatcell{-0.077}{0.000} & --- & \heatcell{-0.011}{1.000} & \heatcell{-0.170}{0.000} & \heatcell{-0.031}{0.813} & \heatcell{-0.183}{0.000} & \heatcell{-0.193}{0.000} & \heatcell{-0.210}{0.000} & \heatcell{0.365}{0.000} & \heatcell{-0.142}{0.000} & \heatcell{-0.113}{0.000} & \heatcell{-0.172}{0.000} & \heatcell{-0.142}{0.000} & \heatcell{-0.147}{0.000} & \heatcell{-0.207}{0.000} & \heatcell{-0.173}{0.000} & \heatcell{-0.198}{0.000} & \heatcell{-0.103}{0.000} & \heatcell{0.371}{0.000}
\\
eng & \heatcell{-0.196}{0.000} & \heatcell{-0.191}{0.000} & \heatcell{-0.187}{0.000} & \heatcell{-0.066}{0.000} & \heatcell{0.011}{1.000} & --- & \heatcell{-0.159}{0.000} & \heatcell{-0.020}{0.998} & \heatcell{-0.172}{0.000} & \heatcell{-0.182}{0.000} & \heatcell{-0.199}{0.000} & \heatcell{0.376}{0.000} & \heatcell{-0.131}{0.000} & \heatcell{-0.102}{0.000} & \heatcell{-0.161}{0.000} & \heatcell{-0.131}{0.000} & \heatcell{-0.136}{0.000} & \heatcell{-0.196}{0.000} & \heatcell{-0.162}{0.000} & \heatcell{-0.187}{0.000} & \heatcell{-0.092}{0.000} & \heatcell{0.382}{0.000}
\\
fin & \heatcell{-0.037}{0.518} & \heatcell{-0.032}{0.768} & \heatcell{-0.029}{0.903} & \heatcell{0.092}{0.000} & \heatcell{0.170}{0.000} & \heatcell{0.159}{0.000} & --- & \heatcell{0.139}{0.000} & \heatcell{-0.013}{1.000} & \heatcell{-0.023}{0.991} & \heatcell{-0.040}{0.343} & \heatcell{0.535}{0.000} & \heatcell{0.028}{0.929} & \heatcell{0.057}{0.009} & \heatcell{-0.003}{1.000} & \heatcell{0.028}{0.928} & \heatcell{0.023}{0.990} & \heatcell{-0.037}{0.483} & \heatcell{-0.004}{1.000} & \heatcell{-0.028}{0.922} & \heatcell{0.067}{0.000} & \heatcell{0.541}{0.000}
\\
fra & \heatcell{-0.175}{0.000} & \heatcell{-0.171}{0.000} & \heatcell{-0.167}{0.000} & \heatcell{-0.046}{0.116} & \heatcell{0.031}{0.813} & \heatcell{0.020}{0.998} & \heatcell{-0.139}{0.000} & --- & \heatcell{-0.152}{0.000} & \heatcell{-0.162}{0.000} & \heatcell{-0.179}{0.000} & \heatcell{0.397}{0.000} & \heatcell{-0.111}{0.000} & \heatcell{-0.082}{0.000} & \heatcell{-0.141}{0.000} & \heatcell{-0.111}{0.000} & \heatcell{-0.116}{0.000} & \heatcell{-0.176}{0.000} & \heatcell{-0.142}{0.000} & \heatcell{-0.167}{0.000} & \heatcell{-0.072}{0.000} & \heatcell{0.402}{0.000}
\\
hin & \heatcell{-0.024}{0.987} & \heatcell{-0.019}{0.999} & \heatcell{-0.016}{1.000} & \heatcell{0.106}{0.000} & \heatcell{0.183}{0.000} & \heatcell{0.172}{0.000} & \heatcell{0.013}{1.000} & \heatcell{0.152}{0.000} & --- & \heatcell{-0.010}{1.000} & \heatcell{-0.027}{0.948} & \heatcell{0.548}{0.000} & \heatcell{0.041}{0.301} & \heatcell{0.070}{0.000} & \heatcell{0.011}{1.000} & \heatcell{0.041}{0.300} & \heatcell{0.036}{0.543} & \heatcell{-0.024}{0.982} & \heatcell{0.010}{1.000} & \heatcell{-0.015}{1.000} & \heatcell{0.080}{0.000} & \heatcell{0.554}{0.000}
\\
hun & \heatcell{-0.014}{1.000} & \heatcell{-0.009}{1.000} & \heatcell{-0.006}{1.000} & \heatcell{0.115}{0.000} & \heatcell{0.193}{0.000} & \heatcell{0.182}{0.000} & \heatcell{0.023}{0.991} & \heatcell{0.162}{0.000} & \heatcell{0.010}{1.000} & --- & \heatcell{-0.017}{1.000} & \heatcell{0.558}{0.000} & \heatcell{0.051}{0.044} & \heatcell{0.080}{0.000} & \heatcell{0.020}{0.998} & \heatcell{0.051}{0.044} & \heatcell{0.046}{0.120} & \heatcell{-0.014}{1.000} & \heatcell{0.019}{0.999} & \heatcell{-0.005}{1.000} & \heatcell{0.090}{0.000} & \heatcell{0.563}{0.000}
\\
ita & \heatcell{0.003}{1.000} & \heatcell{0.008}{1.000} & \heatcell{0.011}{1.000} & \heatcell{0.132}{0.000} & \heatcell{0.210}{0.000} & \heatcell{0.199}{0.000} & \heatcell{0.040}{0.343} & \heatcell{0.179}{0.000} & \heatcell{0.027}{0.948} & \heatcell{0.017}{1.000} & --- & \heatcell{0.575}{0.000} & \heatcell{0.068}{0.000} & \heatcell{0.097}{0.000} & \heatcell{0.037}{0.478} & \heatcell{0.068}{0.000} & \heatcell{0.063}{0.001} & \heatcell{0.003}{1.000} & \heatcell{0.036}{0.533} & \heatcell{0.012}{1.000} & \heatcell{0.107}{0.000} & \heatcell{0.580}{0.000}
\\
jpn & \heatcell{-0.572}{0.000} & \heatcell{-0.567}{0.000} & \heatcell{-0.564}{0.000} & \heatcell{-0.443}{0.000} & \heatcell{-0.365}{0.000} & \heatcell{-0.376}{0.000} & \heatcell{-0.535}{0.000} & \heatcell{-0.397}{0.000} & \heatcell{-0.548}{0.000} & \heatcell{-0.558}{0.000} & \heatcell{-0.575}{0.000} & --- & \heatcell{-0.508}{0.000} & \heatcell{-0.478}{0.000} & \heatcell{-0.538}{0.000} & \heatcell{-0.508}{0.000} & \heatcell{-0.512}{0.000} & \heatcell{-0.573}{0.000} & \heatcell{-0.539}{0.000} & \heatcell{-0.563}{0.000} & \heatcell{-0.468}{0.000} & \heatcell{0.005}{1.000}
\\
kor & \heatcell{-0.064}{0.001} & \heatcell{-0.060}{0.004} & \heatcell{-0.056}{0.010} & \heatcell{0.065}{0.001} & \heatcell{0.142}{0.000} & \heatcell{0.131}{0.000} & \heatcell{-0.028}{0.929} & \heatcell{0.111}{0.000} & \heatcell{-0.041}{0.301} & \heatcell{-0.051}{0.044} & \heatcell{-0.068}{0.000} & \heatcell{0.508}{0.000} & --- & \heatcell{0.029}{0.892} & \heatcell{-0.030}{0.850} & \heatcell{0.000}{1.000} & \heatcell{-0.005}{1.000} & \heatcell{-0.065}{0.001} & \heatcell{-0.031}{0.810} & \heatcell{-0.056}{0.012} & \heatcell{0.039}{0.346} & \heatcell{0.513}{0.000}
\\
nld & \heatcell{-0.094}{0.000} & \heatcell{-0.089}{0.000} & \heatcell{-0.085}{0.000} & \heatcell{0.036}{0.576} & \heatcell{0.113}{0.000} & \heatcell{0.102}{0.000} & \heatcell{-0.057}{0.009} & \heatcell{0.082}{0.000} & \heatcell{-0.070}{0.000} & \heatcell{-0.080}{0.000} & \heatcell{-0.097}{0.000} & \heatcell{0.478}{0.000} & \heatcell{-0.029}{0.892} & --- & \heatcell{-0.059}{0.004} & \heatcell{-0.029}{0.892} & \heatcell{-0.034}{0.692} & \heatcell{-0.094}{0.000} & \heatcell{-0.060}{0.003} & \heatcell{-0.085}{0.000} & \heatcell{0.010}{1.000} & \heatcell{0.484}{0.000}
\\
pol & \heatcell{-0.034}{0.661} & \heatcell{-0.030}{0.874} & \heatcell{-0.026}{0.960} & \heatcell{0.095}{0.000} & \heatcell{0.172}{0.000} & \heatcell{0.161}{0.000} & \heatcell{0.003}{1.000} & \heatcell{0.141}{0.000} & \heatcell{-0.011}{1.000} & \heatcell{-0.020}{0.998} & \heatcell{-0.037}{0.478} & \heatcell{0.538}{0.000} & \heatcell{0.030}{0.850} & \heatcell{0.059}{0.004} & --- & \heatcell{0.030}{0.849} & \heatcell{0.026}{0.966} & \heatcell{-0.035}{0.627} & \heatcell{-0.001}{1.000} & \heatcell{-0.025}{0.969} & \heatcell{0.069}{0.000} & \heatcell{0.543}{0.000}
\\
por & \heatcell{-0.064}{0.001} & \heatcell{-0.060}{0.004} & \heatcell{-0.056}{0.010} & \heatcell{0.065}{0.001} & \heatcell{0.142}{0.000} & \heatcell{0.131}{0.000} & \heatcell{-0.028}{0.928} & \heatcell{0.111}{0.000} & \heatcell{-0.041}{0.300} & \heatcell{-0.051}{0.044} & \heatcell{-0.068}{0.000} & \heatcell{0.508}{0.000} & \heatcell{-0.000}{1.000} & \heatcell{0.029}{0.892} & \heatcell{-0.030}{0.849} & --- & \heatcell{-0.005}{1.000} & \heatcell{-0.065}{0.001} & \heatcell{-0.031}{0.809} & \heatcell{-0.056}{0.012} & \heatcell{0.039}{0.347} & \heatcell{0.513}{0.000}
\\
rus & \heatcell{-0.060}{0.004} & \heatcell{-0.055}{0.013} & \heatcell{-0.052}{0.033} & \heatcell{0.069}{0.000} & \heatcell{0.147}{0.000} & \heatcell{0.136}{0.000} & \heatcell{-0.023}{0.990} & \heatcell{0.116}{0.000} & \heatcell{-0.036}{0.543} & \heatcell{-0.046}{0.120} & \heatcell{-0.063}{0.001} & \heatcell{0.512}{0.000} & \heatcell{0.005}{1.000} & \heatcell{0.034}{0.692} & \heatcell{-0.026}{0.966} & \heatcell{0.005}{1.000} & --- & \heatcell{-0.060}{0.003} & \heatcell{-0.027}{0.951} & \heatcell{-0.051}{0.039} & \heatcell{0.044}{0.160} & \heatcell{0.517}{0.000}
\\
spa & \heatcell{0.001}{1.000} & \heatcell{0.005}{1.000} & \heatcell{0.009}{1.000} & \heatcell{0.130}{0.000} & \heatcell{0.207}{0.000} & \heatcell{0.196}{0.000} & \heatcell{0.037}{0.483} & \heatcell{0.176}{0.000} & \heatcell{0.024}{0.982} & \heatcell{0.014}{1.000} & \heatcell{-0.003}{1.000} & \heatcell{0.573}{0.000} & \heatcell{0.065}{0.001} & \heatcell{0.094}{0.000} & \heatcell{0.035}{0.627} & \heatcell{0.065}{0.001} & \heatcell{0.060}{0.003} & --- & \heatcell{0.034}{0.681} & \heatcell{0.009}{1.000} & \heatcell{0.104}{0.000} & \heatcell{0.578}{0.000}
\\
tur & \heatcell{-0.033}{0.714} & \heatcell{-0.029}{0.905} & \heatcell{-0.025}{0.973} & \heatcell{0.096}{0.000} & \heatcell{0.173}{0.000} & \heatcell{0.162}{0.000} & \heatcell{0.004}{1.000} & \heatcell{0.142}{0.000} & \heatcell{-0.010}{1.000} & \heatcell{-0.019}{0.999} & \heatcell{-0.036}{0.533} & \heatcell{0.539}{0.000} & \heatcell{0.031}{0.810} & \heatcell{0.060}{0.003} & \heatcell{0.001}{1.000} & \heatcell{0.031}{0.809} & \heatcell{0.027}{0.951} & \heatcell{-0.034}{0.681} & --- & \heatcell{-0.024}{0.980} & \heatcell{0.070}{0.000} & \heatcell{0.544}{0.000}
\\
ukr & \heatcell{-0.009}{1.000} & \heatcell{-0.004}{1.000} & \heatcell{-0.001}{1.000} & \heatcell{0.120}{0.000} & \heatcell{0.198}{0.000} & \heatcell{0.187}{0.000} & \heatcell{0.028}{0.922} & \heatcell{0.167}{0.000} & \heatcell{0.015}{1.000} & \heatcell{0.005}{1.000} & \heatcell{-0.012}{1.000} & \heatcell{0.563}{0.000} & \heatcell{0.056}{0.012} & \heatcell{0.085}{0.000} & \heatcell{0.025}{0.969} & \heatcell{0.056}{0.012} & \heatcell{0.051}{0.039} & \heatcell{-0.009}{1.000} & \heatcell{0.024}{0.980} & --- & \heatcell{0.095}{0.000} & \heatcell{0.569}{0.000}
\\
vie & \heatcell{-0.104}{0.000} & \heatcell{-0.099}{0.000} & \heatcell{-0.096}{0.000} & \heatcell{0.026}{0.962} & \heatcell{0.103}{0.000} & \heatcell{0.092}{0.000} & \heatcell{-0.067}{0.000} & \heatcell{0.072}{0.000} & \heatcell{-0.080}{0.000} & \heatcell{-0.090}{0.000} & \heatcell{-0.107}{0.000} & \heatcell{0.468}{0.000} & \heatcell{-0.039}{0.346} & \heatcell{-0.010}{1.000} & \heatcell{-0.069}{0.000} & \heatcell{-0.039}{0.347} & \heatcell{-0.044}{0.160} & \heatcell{-0.104}{0.000} & \heatcell{-0.070}{0.000} & \heatcell{-0.095}{0.000} & --- & \heatcell{0.474}{0.000}
\\
zho & \heatcell{-0.577}{0.000} & \heatcell{-0.573}{0.000} & \heatcell{-0.569}{0.000} & \heatcell{-0.448}{0.000} & \heatcell{-0.371}{0.000} & \heatcell{-0.382}{0.000} & \heatcell{-0.541}{0.000} & \heatcell{-0.402}{0.000} & \heatcell{-0.554}{0.000} & \heatcell{-0.563}{0.000} & \heatcell{-0.580}{0.000} & \heatcell{-0.005}{1.000} & \heatcell{-0.513}{0.000} & \heatcell{-0.484}{0.000} & \heatcell{-0.543}{0.000} & \heatcell{-0.513}{0.000} & \heatcell{-0.517}{0.000} & \heatcell{-0.578}{0.000} & \heatcell{-0.544}{0.000} & \heatcell{-0.569}{0.000} & \heatcell{-0.474}{0.000} & ---
\\
\bottomrule
\end{tabular}
}
\end{subtable} 
\end{minipage}
\hfill
  \begin{minipage}[t!]{0.15\textwidth}
    \centering
    \begin{tikzpicture}[x=1cm,y=1cm]
      \shade[top color=red,bottom color=yellow]
            (0,0) rectangle (0.3,5);
      \draw (0.3,0) -- +(0.4,0) node[right]{0.0};
      \draw (0.3,5) -- +(0.4,0) node[right]{0.65};
    \end{tikzpicture}
    \caption*{\small \(\lvert\Delta\text{compressibility}\rvert\)}%
  \end{minipage}

  \caption{Results of TukeyHSD tests between all languages.  
           Each cell shows the signed difference (bold if \(p<0.05\));  
           shading from yellow→red indicates \(\lvert\Delta\rvert\) on a 0–0.65 scale.}
  \label{tab:tukey-full}

\end{table*}

\end{document}